%% file: ms.tex
\pdfoutput=1 

\documentclass{article}

\newif\ifanonymize 
\anonymizefalse 

\input{neurips_header.tex}
\input{preamble.tex}
\begin{document}

\maketitle


\input{0_abstract.tex}
\input{1_intro.tex}
\input{2_background.tex}
\input{3_method.tex}

\input{4_results.tex}
\input{5_conclusion.tex}

\anonfinal{}{\input{_acknowledgements.tex}}


\bibliographystyle{unsrt}

\anonfinal{\pagebreak}{}
\input{_supp.tex}

\end{document}

%% file: neurips_header.tex
    \usepackage[nonatbib, preprint]{neurips_2019}

\usepackage[utf8]{inputenc} 
\usepackage[T1]{fontenc}    
\usepackage{url}            
\usepackage{booktabs}       
\usepackage{amsfonts}       
\usepackage{nicefrac}       
\usepackage{microtype}      

\title{Disentangled Attribution Curves for Interpreting Random Forests and Boosted Trees}

\author{
    Summer Devlin\\
    EECS Department\\
    UC Berkeley\\
    \texttt{devlins@berkeley.edu}
    \And
    Chandan Singh\\
    EECS Department\\
    UC Berkeley\\
    \texttt{cs1@berkeley.edu}
    \And
    W. James Murdoch\\
    Statistics Department\\
    UC Berkeley\\
    \texttt{jmurdoch@berkeley.edu}
    \And
    Bin Yu\\
    Statistics, EECS Department\\
    UC Berkeley\\
    \texttt{binyu@berkeley.edu}
}

%% file: preamble.tex
\usepackage{graphicx}
\usepackage[usenames, dvipsnames]{xcolor}
\usepackage{footnote}

\usepackage{textgreek} 
\usepackage{amsmath} 
\usepackage{float}
\usepackage{placeins} 

\usepackage{tablefootnote} 
\usepackage[roman]{parnotes} 
\usepackage{tabulary}
\usepackage{tabularx}

\newcommand{\vx}{{\mathbf x}}
\usepackage{changepage}

\usepackage{algorithm,algorithmicx}  
\usepackage[noend]{algpseudocode}
\definecolor{mygray}{gray}{0.5}
\definecolor{cblue}{RGB}{8, 85, 153}
\definecolor{darkblue}{RGB}{1, 43, 112}

\ifanonymize
  \newcommand{\anonfinal}[2]{#1}
\else
  \newcommand{\anonfinal}[2]{#2}
\fi

\newcommand{\CommentInline}[1]{\State \textcolor{mygray}{\# #1}}

\newcommand{\fref}[1]{Fig~\ref{#1}}
\newcommand{\sref}[1]{Sec~\ref{#1}}
\newcommand{\tref}[1]{Table~\ref{#1}}
\newcommand{\aref}[1]{Algorithm~\ref{#1}}

\usepackage{amsfonts}

\usepackage[
  pdfstartpage=30,
  pdfpagemode=UseOutlines,
  bookmarks,
  bookmarksopen,
  pdfstartview=Fit,
  pdfview=Fit,
  colorlinks,
  linktocpage,
  linkcolor=darkblue,
  citecolor=darkblue,
  pagebackref=true,
  hypertexnames=false
]{hyperref}

%% file: 0_abstract.tex
\label{sec:abstract}

\begin{abstract}
  Tree ensembles, such as random forests and AdaBoost, are ubiquitous machine learning models known for achieving strong predictive performance across a wide variety of domains. However, this strong performance comes at the cost of interpretability (i.e. users are unable to understand the relationships a trained random forest has learned and why it is making its predictions). In particular, it is challenging to understand how the contribution of a particular feature, or group of features, varies as their value changes. To address this, we introduce \underline{D}isentangled \underline{A}ttribution \underline{C}urves (DAC), a method to provide interpretations of tree ensemble methods in the form of (multivariate) feature importance curves. For a given variable, or group of variables, DAC plots the importance of a variable(s) as their value changes. We validate DAC on real data by showing that the curves can be used to increase the accuracy of logistic regression while maintaining interpretability, by including DAC as an additional feature. In simulation studies, DAC is shown to out-perform competing methods in the recovery of conditional expectations. Finally, through a case-study on the bike-sharing dataset, we demonstrate the use of DAC to uncover novel insights into a dataset.
  
\end{abstract}

%% file: 1_intro.tex
\section{Introduction}
\label{sec:intro}


Modern machine learning models have demonstrated strong predictive performance across a wide variety of settings. However, these models have become increasingly difficult to interpret, limiting their use in fields such as medicine \cite{litjens2017survey} and policy-making \cite{brennan2013emergence}. Moreover, the use of such models has come under increasing scrutiny as they struggle with issues such as fairness \cite{dwork2012fairness} and regulatory pressure \cite{goodman2016european}. To address these concerns, research in interpretable machine learning has received an increasing amount of attention \cite{murdoch2019interpretable, doshi2017towards}. Interpretable machine learning aims to increase the descriptive accuracy of models (i.e. how well they can be described to a relevant audience) without losing predictive accuracy \cite{murdoch2019interpretable}.

Here, we focus on tree ensembles such as random forests \cite{breiman2001random} and AdaBoost \cite{freund1996experiments}, which are widely used and known to have strong predictive performance. However, their complex nonlinear form and sheer number of parameters has made them difficult to interpret, beyond limited notions such as individual feature importance \cite{breiman2001random, strobl2008conditional, lundberg2018consistent}. 


To address these problems, we introduce \underline{D}isentangled \underline{A}ttribution \underline{C}urves (DAC), a method which yields importance curves for any feature (or group of features) to the predictions made by a tree ensemble model\footnote{Code, notebooks, and models with a simple API for reproducing all results is available \anonfinal{as a zip file (de-anonymized code will be on github)}{at \url{https://github.com/csinva/disentangled_attribution_curves}}.}. These curves provide a simple post-hoc way to interpret how different features interact within a tree ensemble. Through experiments on real and simulated datasets, we find that DAC provides meaningful interpretations, can identify groundtruth importances, and can be used for simple feature engineering to improve predictive accuracy for a linear model.

\begin{figure}[H]
    \centering
    \includegraphics[width=\textwidth]{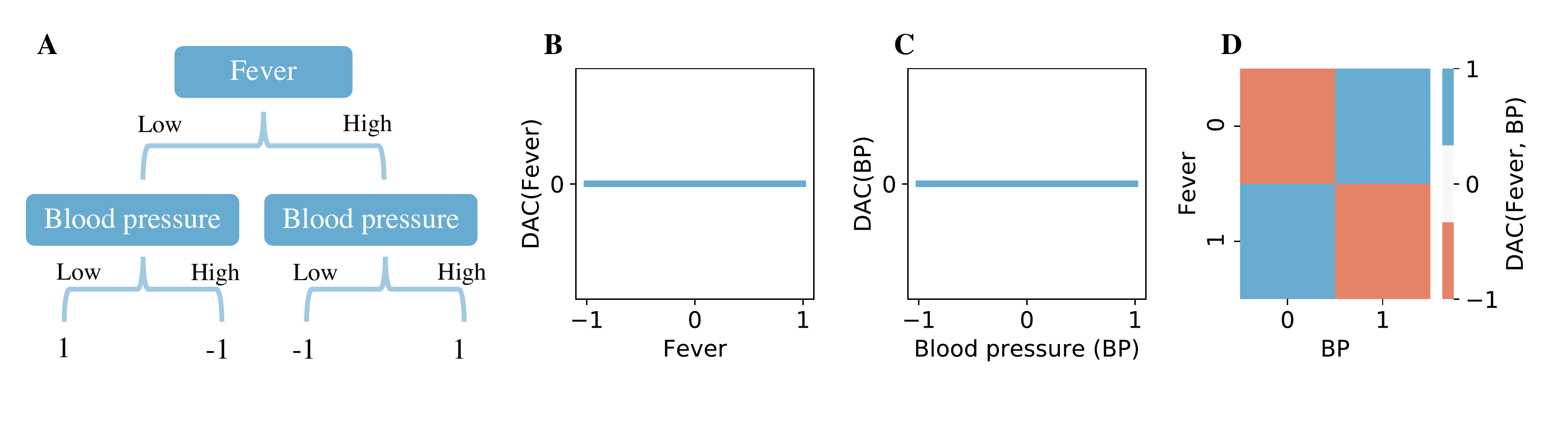}
    \caption{Toy example illustrating DAC. DAC succesfully learns that the interaction between features determines the prediction, not the features in isolation. (A) Tree representing the XOR function. (B, C) DAC single-feature importance for \textit{Fever} and \textit{Blood pressure}, respectively. (D) DAC importance for interaction between \textit{Fever} and \textit{Blood pressure}.}
    \label{fig:intro}
\end{figure}

\paragraph{A simple example} \fref{fig:intro} demonstrates a simple example of how DAC works. The tree in \fref{fig:intro}A depicts a decision tree which performs binary classification using two features (representing the XOR function). In this problem, knowing the value of one of the features without knowledge of the other feature yields no information - the classifier still has a 50\% chance of predicting either class. As a result, DAC produces curves which assign 0 importance to either feature on its own (\fref{fig:intro}B-C). Knowing both features yields perfect information about the classifier, and thus the DAC curve for both features together correctly shows that the interaction of the features produces the model's predictions.

In this instance, due to their inability to describe interactions, common feature-based importance scores would assign non-zero importance to both features \cite{breiman2001random, breiman1984classification, lundberg2018consistent}. Consequently, these scalar feature-importance scores are actually an opaque combination of the importance of the individual feature and that feature's interactions with other features. In contrast, DAC explicitly disentangles the importances of individual features and the interactions between them.

The paper is organized as follows: \sref{sec:background} gives background on related methods, \sref{sec:method} introduces the method, \sref{sec:results} displays results on both simulated and real datasets, and \sref{sec:conclusion} ends with some conclusions.

%% file: 2_background.tex
\section{Background}
\label{sec:background}

This section briefly covers related work and its relation to DAC.

\subsection{Scalar importance scores}

The most common interpretations for tree ensembles take the form of scalar feature-importance scores. They are often useful, but their scalar form restricts them from providing as much information as the curves provided by DAC. Importantly, the importance scores for any feature implicitly contain scores for the interaction of that feature with other features, making it unclear how to interpret the feature importance in isolation.

Impurity-based scores (e.g. mean-decrease impurity) \cite{breiman1984classification, breiman2001random} are widely used in tree ensemble interpretation. These scores are computed by averaging the impurity reduction for all splits where a feature is used. When using the Gini impurity, these scores are sometimes referred to as the Gini importance. 

Another popular score is the permutation importance (i.e. mean decrease accuracy) \cite{breiman2001random}. It is defined as the mean decrease in classification accuracy after permuting a feature over out-of-bag samples. More recently, this method has been improved to develop conditional variable importance via a conditional permutation scheme \cite{strobl2008conditional}.

Besides these scores, there are some model-agnostic methods to compute feature importance at the prediction-level (i.e. for one specific prediction) \cite{ribeiro2016should}. One popular example is Shap values, which provide prediction-level scores for individual variables \cite{lundberg2017unified, lundberg2018consistent}.

\paragraph{Interaction importance scores}

Moving beyond single-variable importance scores, some more recent methods look for interactions between different variables.

For example, recent work \cite{basu2018iterative, kumbier2018refining} stabilizes the training of random forests and then extracts interactions by looking at co-occurrences of features along decision paths. Additionally, some work has begun to address looking at interactions at the prediction-level for random forests (e.g. \cite{lundberg2018consistent}). Recent methods in the deep learning literature follow a similar intuition of explicitly computing importances for feature interactions \cite{singh2018hierarchical, murdoch2018beyond, tsang2017detecting}. There also exist other methods for calculating interaction importance which are model-agnostic \cite{friedman2008predictive, molnar2019quantifying}.

\subsection{Importance curves}

Some recent interpretation methods compute curves to convey importances for different features as their value changes. However, these previous methods operate solely on the inputs and outputs of a model, ignoring the rich information contained within the model. 

Let $\hat{f}$ be the fitted model of interest, $X$ the input features, $S$ the set of the variables of interest, and $C$ the set of remaining variables. Partial dependence plots (PDP) simply plot the expectation of the marginal function on $X_S$ \cite{friedman2001greedy}, which amounts to integrating out the values of $X_C$: $\hat{f}(X_S) = \mathbb E _ {X_C} \hat f (X_S, X_C)$. This assumes that the features in $S$ and $C$ are independent, otherwise the points being averaged over do not accurately represent the underlying data distribution. Independence is often an unrealistic assumption which can cause serious issues for this method \cite{strobl2008conditional}. Individual conditional expectation (ICE) plots \cite{goldstein2015peeking} are very similar to PDP (but yield one curve for each data point), and suffer from the same independence assumption.

Aggregated local effects (ALE) plots calculate local differences in predictions based on the variable of interest, conditioned on $x_C$ \cite{apley2016visualizing}. Then these local effects are aggregated from some baseline. These local effects ameliorate the issue with assuming independent variables, but introduce sensitive hyperparameters, such as a window size and a baseline, which can alter the interpretation. DAC helps fix the need for choosing these hyperparameters by leveraging the internals of the random forest.



%% file: 3_method.tex
\section{Disentangled Attribution Curves}
\label{sec:method}

DAC yields post-hoc interpretations for tree ensembles which have already been trained. At a high level, DAC importance(s) for a feature (or a group of features) correspond to a fitted model's average prediction for training data points classified using only that feature (or group of features). This procedure ensures that the importances for individual features and their interactions are disentangled, since the DAC importance for individual features completely ignores other features. 


DAC also satisfies the so-called \textit{missingness} property \cite{lundberg2017unified}: features which are not used in the forest do not contribute to the importance for any predictions. This is trivially satisfied, as only features which are used by the forest can change the DAC importance.

Intuitively, DAC is similar to a fairly natural interpretation: the conditional expectation of the model's predictions over the dataset conditioned on particular values of a feature. However, DAC is superior to the simple conditional expectation curve, as it is more faithful to the model's underlying process. For example, consider a model which places high importance on only one of two correlated features. Conditional expectation will assign both features a high importance, whereas DAC will successfully learn to assign high importance to only the appropriate feature.

In this paper, we focus on cases where $S$ contains either one or two variables, yielding importance curves or heatmaps, due to the difficulties associated with visualizing more than 3 variables. However, DAC importances can still be used to investigate higher-order interactions between variables.

DAC curves are constructed in a hierarchical fashion, mirroring the structure of a tree ensemble model. In particular, a tree ensemble is a combination of trees, and each tree is a combination of leaves. Given a trained random forest or AdaBoost model, we first define DAC for individual leaves, then demonstrate how to aggregate from leaves to trees, and trees to forests.

\subsection{Attribution on leaves}
\label{subsec:leaves}
At each leaf node $L$ on a tree $T$, a decision tree makes its prediction by computing the average output value $Y_i$ over a subset $R_L$ of the input data. Without loss of generality, assume this subset is computed by applying a sequence of decision rules of the form $X_p \geq c$. Formally, for a path of length $N$, this corresponds to the region $R$ defined below, resulting in estimate $\hat{Y}$.

\begin{align}
    R_L & = \bigcap_{j=1}^N \{X_{p_j} \geq c_j\} \\
    \hat{Y} & = \frac{1}{|R_L|}\sum_{i: X_i \in R_L} Y_i
\end{align}

Note that, to compute the region $R_L$ requires using information about the set of variables $\{X_{p_j}\}_{j=1}^N$. To compute our DAC estimate for a set of variables $S$, we restrict $R_L$ to $R_{L, S}$ by only using splits for variables which are contained in $S$. In addition, we restrict our consideration to data points $X$ which are close to the center of the leaf. In particular, we require that $|x_j - \mu_{L, j}| \leq k \sigma_{L, j}$, for all $j \in S$, where $k$ is a smoothing hyperparameter fixed to one throughout this paper, and $\mu_{L, j}, \sigma_{L, j}$ denote the mean and standard deviation of $X_j$ contained in $R_L$ (if $|R_L|=1$, we set $\sigma_{L,j} = 0$). This yields our DAC estimate $\hat{Y}_{DAC}(S; L, T)$ for a single leaf:

\begin{align}
    C_L & = \{x \quad \vert \quad |x_j - \mu_{L, j}| \leq  k \sigma_{L, j}\  \forall j \in S\} \\
    R_{S, L} & = \left(\bigcap_{j=1, p_j \in S}^N \{X_{p_j} \geq c_j\} \right) \cap C_L \\
    \hat{Y}_{DAC}(S; L, T) & = \frac{1}{|R_{S,L}|}\sum_{i:X_i \in R_{S,L}} Y_i
\end{align}

As $\hat{Y}_{DAC}(S; L, T)$ is computed using only splits involving variables contained in $S$, we interpret it as the importance of those variables for making the prediction at this particular leaf. 

\subsection{Aggregating leaf attributions to trees and forests}
\label{subsec:leaves_to_trees}

Given the estimates $\hat{Y}_{DAC}(S; L, T)$ for the contribution of a set of variables $S$ to the prediction made at an individual leaf $L$ on tree $T$, we now describe how to aggregate those contributions in order to produce an importance curve for a decision tree. When computing the DAC contribution for a particular point $x \in \mathbb{R}^{|S|}$, we take an average over the DAC contributions of leaves $L$ which are ``nearby'' to $x$. To define proximity, we use the same regions $C_L$ as with leaves. Given these regions, we formally define the $DAC$ importance for a given tree at $x$ to be the weighted average of regions $C_L$ containing $x$. We weight the average by the size of the regions $|C_L|$ in order to account for the size of the leaves in our calculation, and avoid giving undue influence to smaller leaves.

\begin{align}
    f_{DAC}(x; S, T) & = \frac{1}{\sum_{L \text{ in tree}, x \in C_L} |C_L|}\underset{L \text{ in tree}, x \in C_L}{\sum} |C_L| \cdot \hat{Y}_{DAC}(S; L, T)
\end{align}

Finally, given an ensemble of trees $T_1,..., T_M$, with corresponding weights $w_1,...,w_M$, we simply define the attribution curve for the tree ensemble to be the weighted average of the attributions for the individual trees $f_{DAC}(x; S, T_i)$. Note that this mirrors the prediction process for the model, as the predictions for an ensemble of trees are computed by averaging the predictions of it's constituent trees. In particular, when the ensemble is a random forest, $w_i = \frac{1}{M}$.

\begin{align}
    f_{DAC}(x; S) = \sum_{i=1}^M w_i f_{DAC}(x; S, T_i)
\end{align}

\subsection{An algorithm for computing DAC curves}

\aref{algo:dac_curve} presents pseudocode for computing the tree-level DAC curves $f_{DAC}(x; S, T)$ introduced in \sref{subsec:leaves} and \sref{subsec:leaves_to_trees}. A simple analysis of the algorithm shows that its time complexity is $O(n \cdot |S|)$, where $n$ is the number of data points in the training set and $|S|$ is the cardinality of the set of features of interest. For a forest, this complexity grows linearly with the number of trees (although they can be very easily parallelized by computing the curve for trees in parallel). In fact, both for-loops in \aref{algo:dac_curve} can be fully computed in parallel. Similarly, the space complexity of this algorithm is $O(n \cdot |S|)$. 

The above analysis assumes summation over $dac$ and $counts$ (the last lines in the for-loop of \aref{algo:dac_curve} happen in constant time. In practice, the size of these arrays is determined by a pre-specified grid of $X$ values for which the user would like to calculate DAC importances (used to find the size of $dac$ and $counts$). With extremely large grids (beyond what is reasonable in practice), the complexity of the algorithm would depend linearly on the sizes of the arrays $dac$ and $counts$. The above analysis also assumes that the total time complexity for finding the subset of points in each leaf which obey the rules for that leaf across all leaves is $O(n)$.

\begin{algorithm}[H]
    \caption{DAC Importance curve generation for a tree}
    \label{algo:dac_curve}
    \small
    \textbf{DAC}(a tree $T, X, y$, set of features $S$, smoothing parameter $k$)
    \begin{algorithmic}
    \State initialize importance array $dac$ with zero values
    \State initialize counts array $counts$ with zero values
    \For{\text{each leaf $l$ in $T$}}
        \vspace{3pt}
        \CommentInline{Find the points in a leaf which use rules in $S$}
        \State Let $R$ be the set of rules encountered on the path from the root to $l$ concerning features in $S$
        \State Let $X'$ be the subset of points in $X$ such that for each point $x$, $x$ complies with all rules in $R$
        \State Let $y'$ contain the corresponding $y$ values
        \\
        \CommentInline{Ensure feature values in a leaf are not outliers}
        \For{\text{each feature $i$ in $S$}}
        \State Compute the mean and standard deviation of $X'_i$, denoted $\mu_{xi}$, $\sigma_{xi}$
        \State Remove row $j$ in $X'$ and $y'$ if $X'_j \notin \mu_{xi} \pm k * \sigma_{xi}$
        \EndFor
        \vspace{6pt}
        \CommentInline{Add the contribution of this leaf to the DAC curve}
        \State Define $\mu_{y}$ and $c_{y}$ to be the mean and cardinality of $y'$ respectively
        \State Define interval $int$ centered at $(\mu_{xi}, ..., \mu_{xj})$ and extending by $\pm (\sigma_{xi}, ..., \sigma_{xj})$
        \State $dac = dac +  \mu_{y} \cdot c_{y}$ over interval $int$
        \State $counts = counts + c_{y}$ to  over interval $int$
    \EndFor
        \item \textbf{return} $c$ / $counts$ (normalize elementwise)
    \end{algorithmic}
\end{algorithm}


%% file: 4_results.tex
\section{Results}
\label{sec:results}

We now validate, both quantitatively and qualitatively, our introduced DAC importances. First, \sref{subsec:quantitative_real_data} demonstrates that DAC curves can be used as additional features in a linear model to help close the predictive gap between linear models and random forests. Next, through simulation studies, \sref{subsec:simulation} quantifies that DAC curves provide a better approximation to conditional expectation than other approaches. Finally, \sref{subsec:qualitative} shows a case study on the bike-sharing dataset, revealing insights into the data. 


\subsection{Automated feature engineering with DAC}
\label{subsec:quantitative_real_data}
We now show that single-feature DAC curves produced by \aref{algo:dac_curve} can be used as additional features to increase the accuracy of a logistic regression model on real-world datasets. This increase in accuracy indicates that DAC curves are capturing meaningful relationships in the data, which are predictive of the outcome in question.

To compute these results, we use numerous classification datasets retrieved from the Penn Machine Learning Benchmarks \cite{olson2017pmlb}. The datasets include a variety of data types such as classifying liver disorders (the \textit{bupa} dataset \cite{mcdermott2016diagnosing}) or evaluating the condition of cars (the \textit{car} dataset \cite{bohanec1988knowledge}). To find datasets containing significant non-linear effects, we used only datasets where random forests outperformed logistic regression by at least 5\% (\fref{fig:pmlb_classification} shows results for datasets where random forests outperform logistic regression by any positive margin).

We randomly partition each of the selected datasets into a training set (75\% of the data) and a testing set (25\% of the data). We begin by training a random forest model on the training set\footnote{all tree ensembles in this work are trained using scikit-learn \cite{pedregosa2011scikit} and contain fifty trees.}. Then, the fitted model and the training data are used to identify the feature with the highest Gini importance, and to compute the corresponding single-feature DAC curve.

We then use the computed DAC curve to define a new feature for use by a logistic regression model. The curve is a univariate function which transforms a single feature into its DAC importance. Therefore, for each data point $X$, we append the value of the DAC curve at $X_i$, where $i$ is the important feature. On the training set, we fit a logistic regression model using both the original features and the new feature constructed above. Finally, we test the fitted model on the testing set, where we we use the same DAC curve to append the single-feature DAC feature on the testing set \footnote{Importantly, the DAC curve is constructed on the training set and never recieves any information about the labels on the testing set.}.

\tref{tab:logit_append} gives results for the accuracy on the testing set. In these instances, adding a single DAC curve feature increased the prediction accuracy, with increases ranging from 3.9\% to 23\%. The increased accuracy reduces the predictive gap between logistic regression and random forests. In one instance (the \textit{bupa} dataset), logistic regression with a single DAC curve even outperforms a random forest model (78.2\% accuracy versus 75.9\% accuracy for the random forest). Classification results for more datasets are given in \fref{fig:pmlb_classification}.

While we only append a single 1-dimensional curve here, this process of feature-engineering can be used to append many DAC curves for different features, and even multi-dimensional DAC curves to model interactions between features.



\begin{table}[H]
    \small

  \label{tab:logit_append}
\centering
\caption{Accuracy results on various datasets. Logistic regression trained on data with one DAC curve as an extra feature outperforms logistic regression on the original features.}
\begin{tabulary}{\textwidth}{LCCCCCCCCCC} 
    \toprule

    	& \textit{monk3} & \textit{irish} & \textit{monk1} & \textit{hayes-roth} & \textit{bupa} & \textit{tokyo1} & \textit{tic-tac-toe} & \textit{buggyCrx} & \textit{agaricus-lepiota} & \textit{car} \\ 
    \midrule
    RF Accuracy & 0.971 & 1.000 & 0.986 & 0.825 & 0.759 & 0.900 & 0.950 & 0.884 & 1.000 & 0.977 \\ 
    Logistic Accuracy & 0.734 & 0.776 & 0.583 & 0.400 & 0.690 & 0.792 & 0.650 & 0.740 & 0.946 & 0.650 \\ 
    \textbf{Logistic Accuracy + DAC} & 0.964 & 1.000 & 0.719 & 0.525 & 0.782 & 0.875 & 0.721 & 0.809 & 0.989 & 0.690 \\ 
    Difference & 0.230 & 0.224 & 0.137 & 0.125 & 0.092 & 0.083 & 0.071 & 0.069 & 0.042 & 0.039 \\ 
    
    \bottomrule
    \end{tabulary}
      
\end{table}

\subsection{Simulation results}
\label{subsec:simulation}

We now use simulations to test how well DAC can recover interactions contained in data. To do so, we generate synthetic data with complex interactions. The functional form of these simulations is in \tref{tab:sim_funcs}, based on simulations used in previous work \cite{tsang2017detecting}.

\begin{table}[H]
  \small
  \caption{Simulation functions.}
  \label{tab:sim_funcs}
  \centering
  \begin{tabular}{lc}
    \toprule
     $F_1(\vx)$\ & $\begin{aligned}[t] 
    &\pi^{x_{1}x_{2}}\sqrt{2x_{3}} - \sin^{-1}(0.5 x_{4}) + \log(|x_3+x_5|+1)
    \end{aligned}$  \\\hline
    
     $F_2(\vx)$ & $\begin{aligned}[t] 
    &  \pi^{x_1x_2}\sqrt{2|x_3|} -\sin^{-1}(0.5x_4) + \log(|x_3+x_5|+1) - x_2 x5
    \end{aligned}$\\ \hline

     $F_3(\vx)$\ & $\begin{aligned}[t] 
    &\exp|x_1-x_3| + |x_2x_3| - x_3^{2|x_4|} + \log(x_4^2+x_5^2)
    \end{aligned}$  \\\hline
    
    $F_4(\vx)$\ & $\begin{aligned}[t] 
    &\exp{|x_{1} - x_{3}|} + |x_{2}x_{3}| - x_{3}^{2|x_{4}|} + (x_{1}x_{4})^{2} + \log(x_{4}^{2} + x_{5}^{2})
    \end{aligned}$  \\\hline
    
     $F_5(\vx)$ & $\begin{aligned}[t] 
    & \frac{1}{1+x_{1}^{2}+x_{2}^{2}+x_{3}^{2}} + \sqrt{\exp(x_{4}+x_{5})}
    \end{aligned}$ \\ \hline
    
     $F_6(\vx)$ & $\begin{aligned}[t] 
    &\exp{(|x_2x_3|+1)} -\exp(|x_3+x_4|+1) + \cos(x_5)
    \end{aligned}$ \\ \hline
    
    $F_7(\vx)$  & $\begin{aligned}[t] 
    &(\arctan(x_1)+\arctan(x_2))^2 +\max(x_3x_4+x_6,0)-\frac{1}{1+(x_4x_5)^2} + \sum_{i=1}^{5}{x_i}
    \end{aligned}$ \\ \hline
    
     $F_8(\vx)$ & $\begin{aligned}[t] 
    &x_1x_2 + 2^{x_3+x_5} + 2^{x_3+x_4+x_5}
    \end{aligned}$  \\ \hline
    
     $F_{9}(\vx)$ & $\begin{aligned}[t] 
    &\arctan(x_1x_2+x_3x_4)\sqrt{|x_5|}+\exp(x_5+x_1)
    \end{aligned}$  \\ \hline
    
     $F_{10}(\vx)$ & $\begin{aligned}[t] 
    &\sinh{(x_1+x_2)}+\arccos\left(\arctan(x_3+x_5)\right) + \cos(x_4+x_5)
    \end{aligned}$  \\
    \bottomrule
    
  \end{tabular}
\end{table}

For each of the functions in \fref{tab:sim_funcs}, we generate a training dataset of 70,000 points and a testing dataset of 15 million points. In all cases, data is generated from a 5-dimensional multivariate Gaussian centered at zero, and points lying outside the interval (-2, 2) are rejected. The data is drawn using different covariance matrices: ``IID'' refers to the features being independent, ``Highly correlated'' refers to the features being highly correlated (covariance matrix has eigenvalues 2, 2, 1, 0, 0), and ``Correlated'' is in between (for specifics, see \fref{fig:covs}).

We then fit a tree ensemble regressor to the generated training data. From these fitted models, we extract DAC/PDP curves using the training data\footnote{PDP curves were calculated using the PDPbox library available at \url{https://github.com/SauceCat/PDPbox}}. Finally, as a proxy for learning the groundtruth interactions, we measure the MSE between these curves and conditional expectation curves calculated on the very large testing set. 


\tref{tab:sim_results} shows the matches between the DAC and PDP curves with the conditional expectation curves. DAC (the left column) matches the conditional expectation curves better than PDP (the right column) by a substantial margin for different data distributions. This is true even when the features are independent. Comparisons between rows are not meaningful, as they are evaluated with different models and data distributions.

\begin{table}[H]
  \caption{Simulation results (averaged over the 5 variables in each of the ten functions in \tref{tab:sim_funcs}). Error is standard error of the mean.}
  \label{tab:sim_results}
  \centering
  \begin{tabular}{ccc}
    \toprule

    & \textbf{DAC MSE} & PDP MSE \\ 
    \midrule
    
    RF IID & 0.776 $\pm$ 0.110 &     1.037 $\pm$ 0.147 \\
    RF correlated & 0.325 $\pm$ 0.052 &     0.477 $\pm$ 0.076 \\
    RF highly correlated & 0.553 $\pm$ 0.081 &     1.459 $\pm$ 0.213 \\
    Adaboost IID & 1.144 $\pm$ 0.162 &     13.688 $\pm$ 1.936 \\
    Adaboost correlated & 1.301 $\pm$ 0.184 &     21.300 $\pm$ 3.012 \\
    Adaboost highly correlated & 2.569 $\pm$ 0.115 &     12.895 $\pm$ 0.577 \\
    \bottomrule
  \end{tabular}
\end{table}

\subsection{Qualitative results}
\label{subsec:qualitative}

We now use our methods in a case study to show how DAC can provide insights into data. We study the Bike-sharing dataset, a popular dataset in interpretability research \cite{molnar2018interpretable}, made open by the company Capital-Bikeshare with additional features added in a later study \cite{fanaee2014event}. The regression task in the dataset is to predict the hourly count of rented bicycles using a number of features such as the temperature, the time of day, and whether it is a holiday. We fit a random forest to the data and then visualize single-feature and multi-feature DAC curves to understand the interactions it has learned.

\begin{figure}[H]
    \centering
    \includegraphics[width=0.4\textwidth]{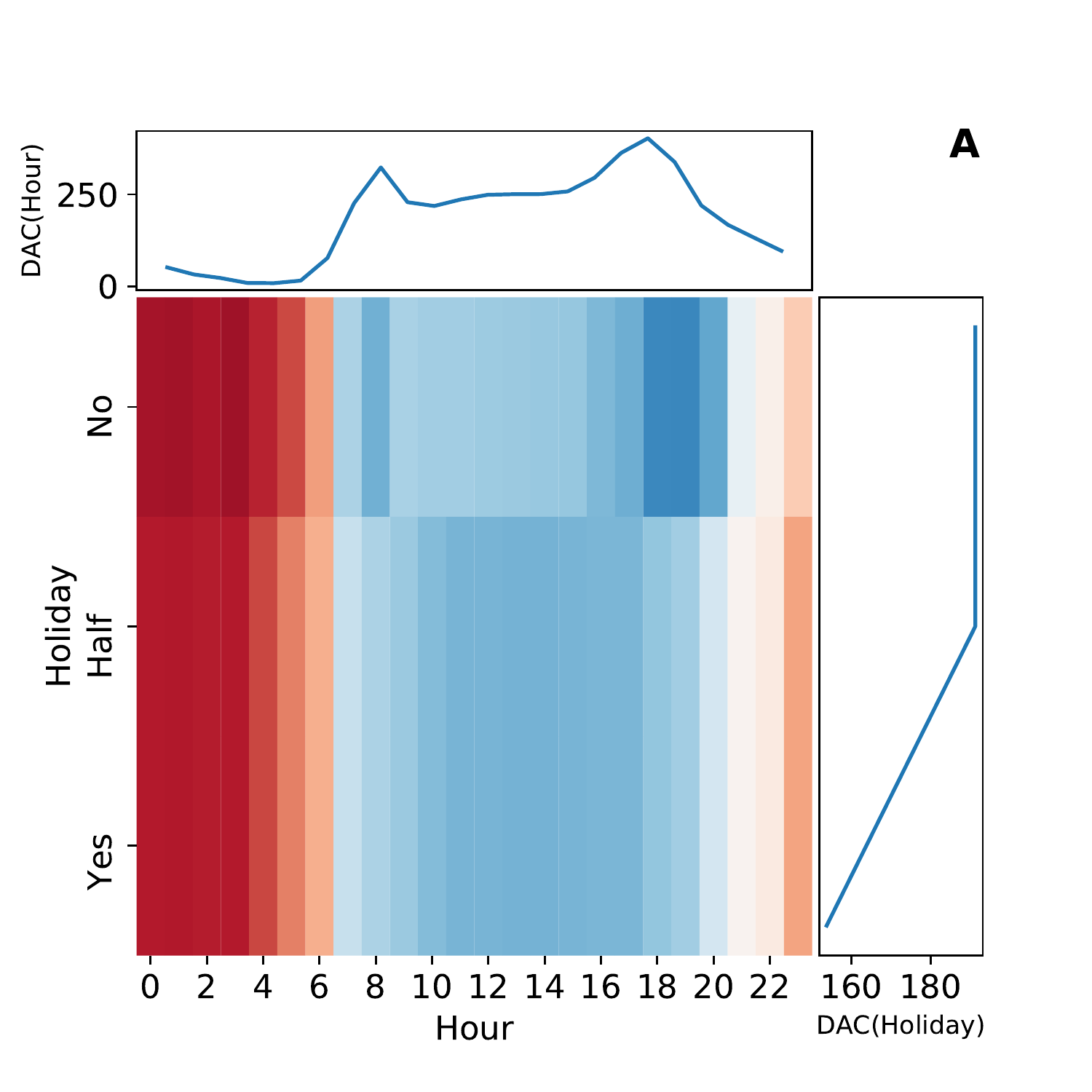}
    \includegraphics[width=0.4\textwidth]{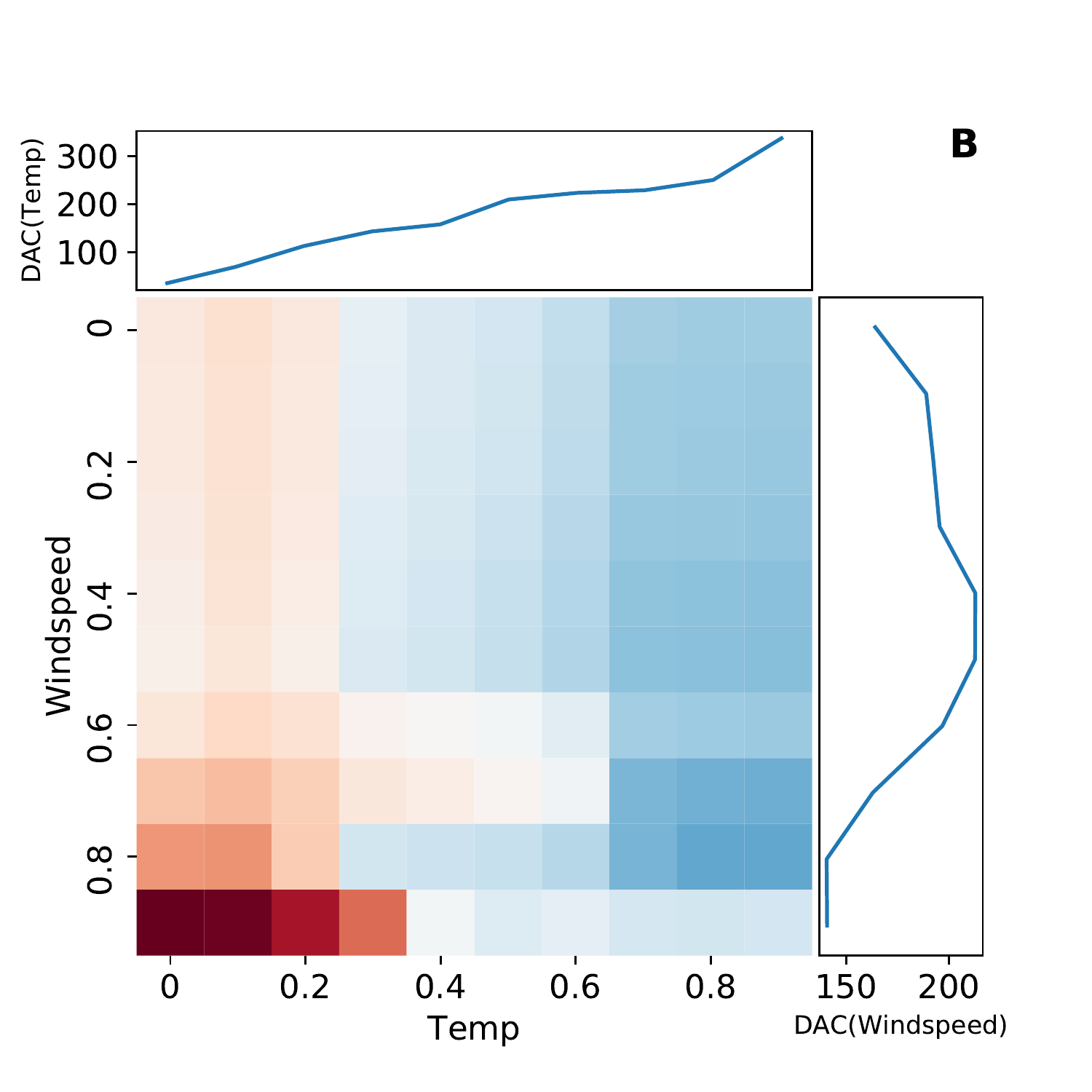}
    \includegraphics[height=2in, width=0.5in]{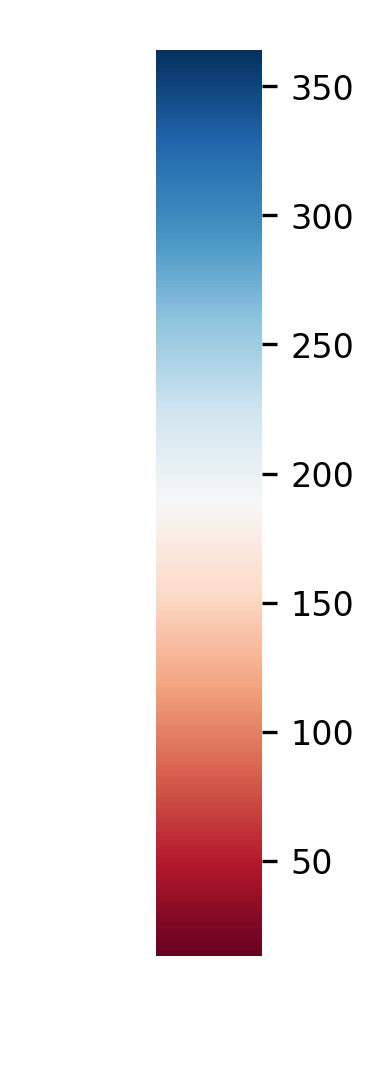}
    \caption{DAC importances and interactions for a random forest fit to the Bike-sharing dataset. Heatmaps show 2-dimensional DAC curves while side plots show 1-dimensional DAC curves.}
    \label{fig:qualitative_main}
\end{figure}

\fref{fig:qualitative_main} shows DAC curves for particular interactions within the data. The heatmaps show 2-dimensional DAC curves (the mean number of rented bicycles in this dataset is 188.7, corresponding to white on the heatmaps). The individual curves on the sides of the heatmaps show 1-dimensional DAC curves for single-feature importance.

\fref{fig:qualitative_main}A shows a qualitatively interesting interaction between the features \textit{Hour} and \textit{Holiday}. Within the single-variable context, the \textit{Hour} feature shows two peaks, corresponding to peak commute times during the day. The curve for \textit{Holiday} shows that people rent fewer bikes during the holidays (note that this is a binary feature, but we can still evaluate the importance of it at non-binary values based on the forest's split points). Interestingly, during holidays, the importance of the time of day decreases at peak commuter time, and is now highest in the middle of the day. This makes intuitive sense, as holidays specifically remove commuter traffic.

\fref{fig:qualitative_main}B shows the interaction between \textit{Temp} and \textit{Windspeed}. The 1-dimensional \textit{Temp} curve has a very strong importance for the random forest's predictions, with higher temperatures resulting in higher model predictions. \textit{Windspeed} has a relatively smaller effect (note the scale difference for the 1-dimensional curves), and a moderate wind results in the highest predictions. The 2-dimensional DAC curve is largely a superposition of these two curves, suggesting little interaction, although at times there are traces of small interactions between wind speed and temperature. \fref{fig:qualitative_supp} shows more DAC curves for strong interactions between features contained in this model.

%% file: 5_conclusion.tex
\section{Conclusion}
\label{sec:conclusion}

In this paper, we have introduced DAC interpretations for tree ensembles, such as random forests and AdaBoost. For a given feature, or pair of features, DAC produces an importance curve, or heat map, which displays the importance of feature(s) as their value changes. This curve intuitively represents local averages of the model's predictions based only on the features(s) of interest. 

Through an extensive validation, we show that DAC curves can be used as features in a logistic regression model to increase its generalization accuracy. Furthermore, in a simulation study, we show that DAC provides a better approximation to the conditional expectation than competing methods. Finally, through a case study on the bike sharing dataset, we show how DAC can be used to extract insights from real data.

This work opens many potentially promising avenues for future work, such as the use of DAC to automatically extract features (perhaps using higher-order interactions), in order to close the gap between logistic regression and random forests while preserving interpretability. Moreover, we think that ideas used in DAC can be extended to a range of other complex models, such as neural networks. Finally, DAC can be used to automatically explore and identify higher-order interactions then the simple interactions analyzed here, and give further insight into how a tree ensemble is making its predictions.

%% file: _acknowledgements.tex
\section*{Acknowledgements}
This research was supported in part by grants ARO W911NF1710005, ONR N00014-16-1-2664, NSF DMS-1613002, and NSF IIS 1741340, an NSERC PGS D fellowship, and an Adobe research award. We thank the Center for Science of Information (CSoI), a US NSF Science and Technology Center, under grant agreement CCF-0939370.

%% file: _supp.tex
\setcounter{table}{0}
\setcounter{figure}{0}
\setcounter{section}{0}
\renewcommand{\thetable}{S\arabic{table}}
\renewcommand{\thefigure}{S\arabic{figure}}
\renewcommand{\thesection}{S\arabic{section}} 

\setlength\floatsep{-200pt}
\setlength\textfloatsep{0\baselineskip}
\setlength\intextsep{0\baselineskip}

\begin{center}
    \section*{Supplement}    
\end{center}


\subsection*{Classification results extended}

\begin{figure}[H]
    \centering
    \includegraphics[width=0.5\textwidth]{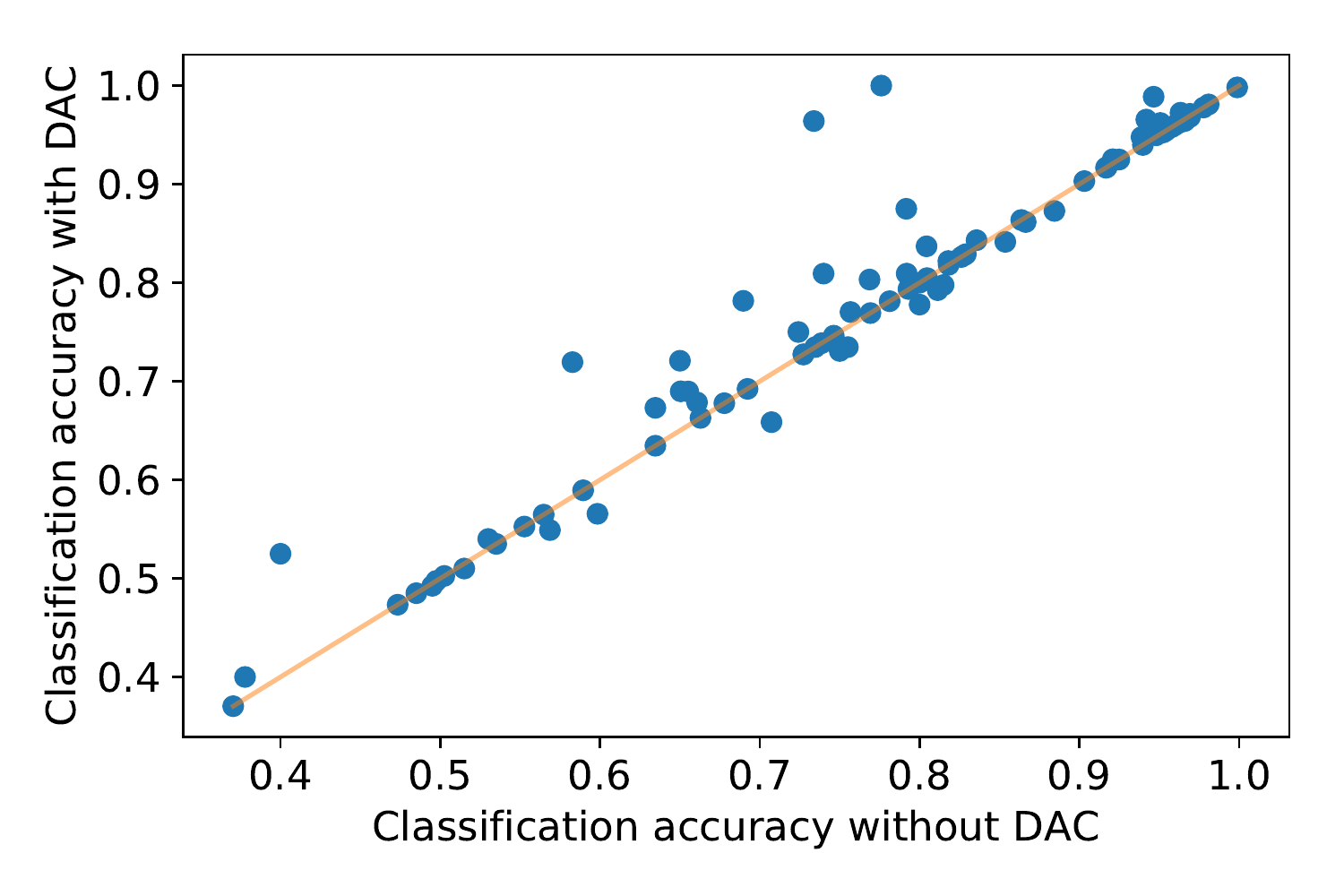}
    \caption{Logistic regression accuracy improves with and without DAC, similiar to \tref{tab:logit_append}. Each point corresponds to classification accuracy for a different dataset. These datasets are the datasets in PMLB \cite{olson2017pmlb} for which a random forest outperforms logistic regression.}
    \label{fig:pmlb_classification}
\end{figure}

\subsection*{Qualitative interactions extended}

\begin{figure}[H]
    \centering
    \includegraphics[width=0.4\textwidth]{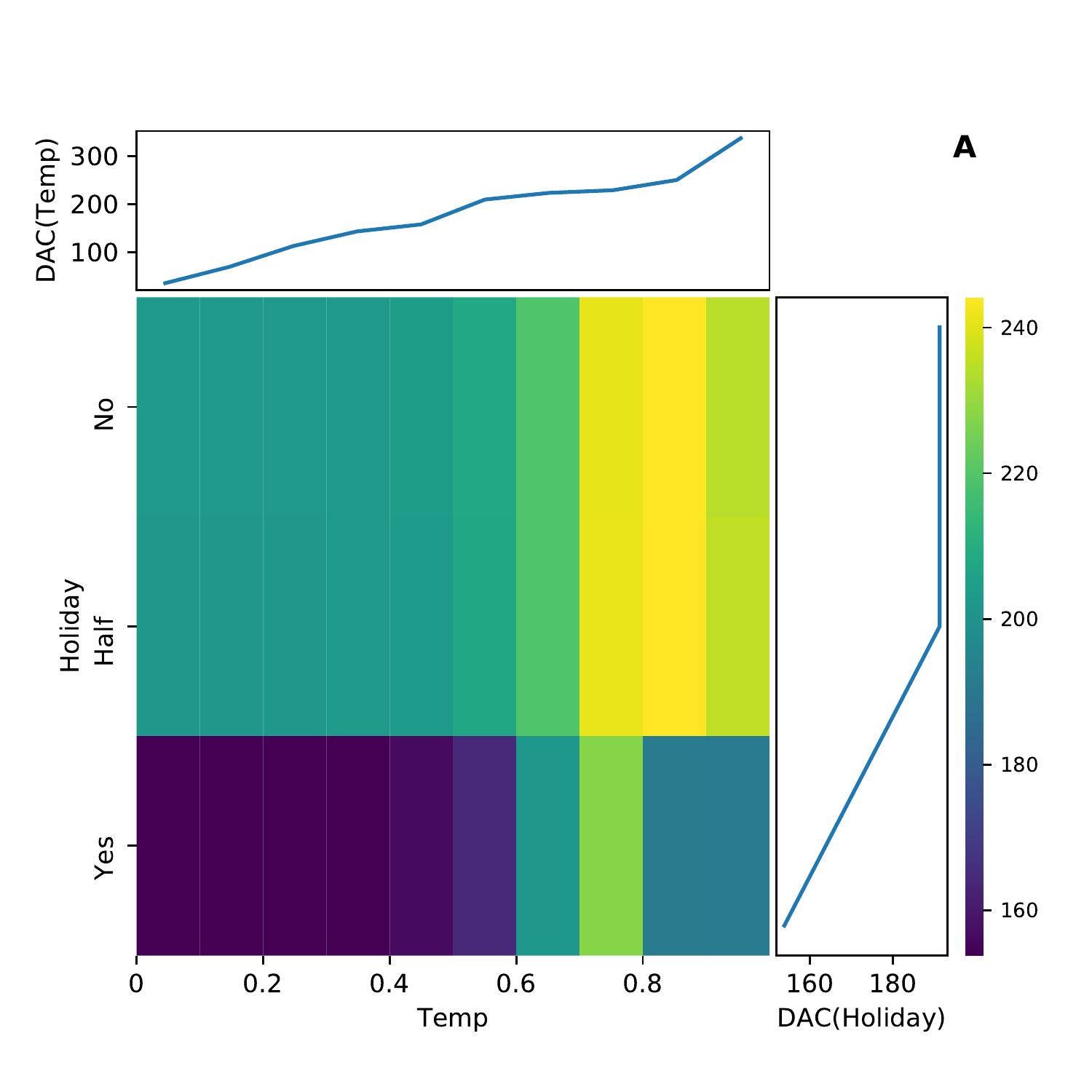}
    \includegraphics[width=0.4\textwidth]{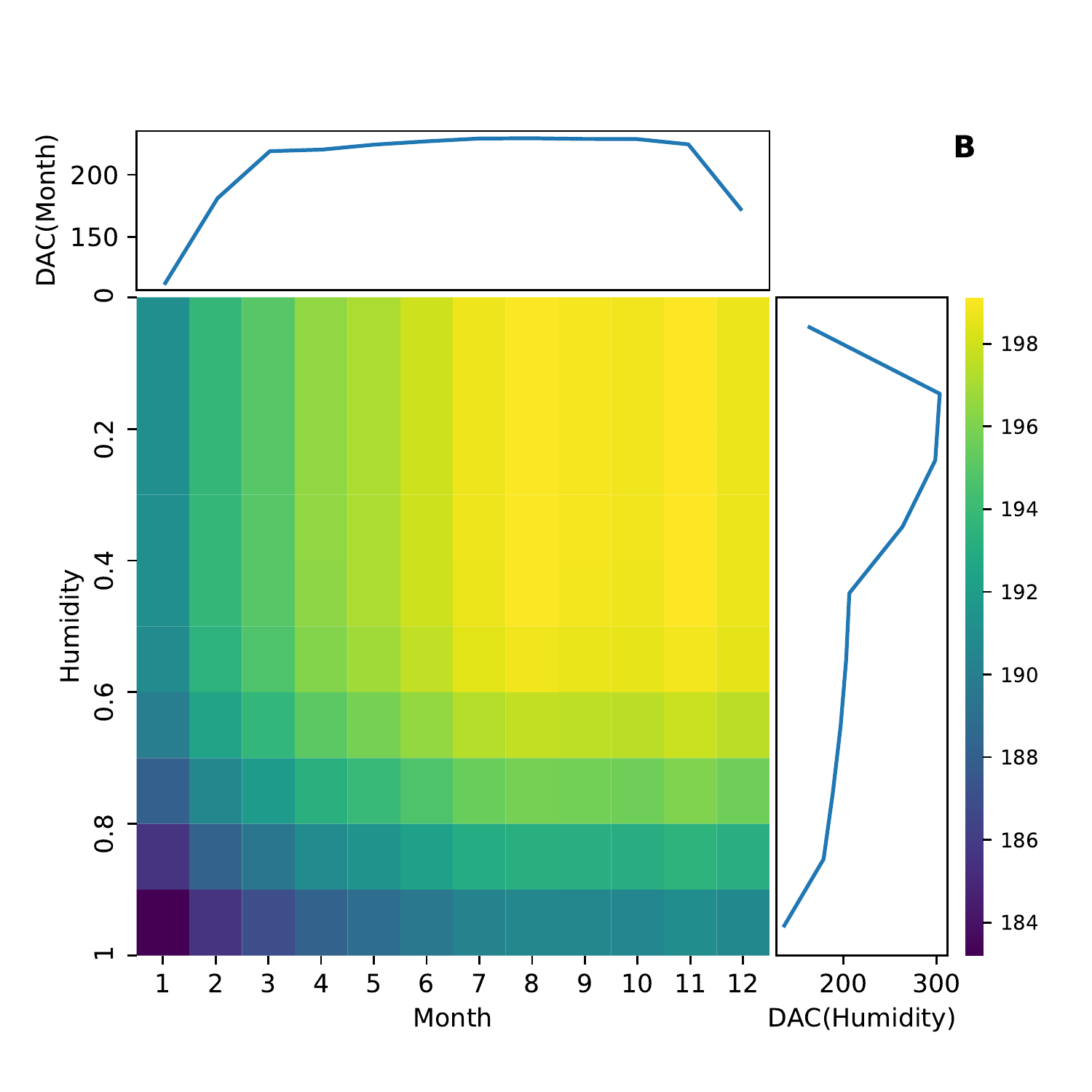}
    \caption{DAC importances and interactions in the Bike dataset. }
    \label{fig:qualitative_supp}
\end{figure}

\fref{fig:qualitative_supp}A shows the interaction of the features \textit{temp} with {holiday}. The peak temperature to rent bikes is decidedly shifted during holidays, indicating that people only bike during very fair weather on their holidays

\fref{fig:qualitative_supp}B shows the interaction between \textit{Month} and \textit{Humidity}. On its own, month is a feature with a very simple effect that could be equally described by season: people rent fewer bikes during the winter months.  Adding humidity broadens the peak month set, and adds more variation to the score on popular and unpopular months.  It is clear that people most prefer non-humid summer and fall days to bike, but that biking on a humid hot day is only a slightly better than biking on a humid cold day.

\subsection*{Simulation settings extended}

\begin{figure}[H]
    \centering
    \includegraphics[width=0.7\textwidth]{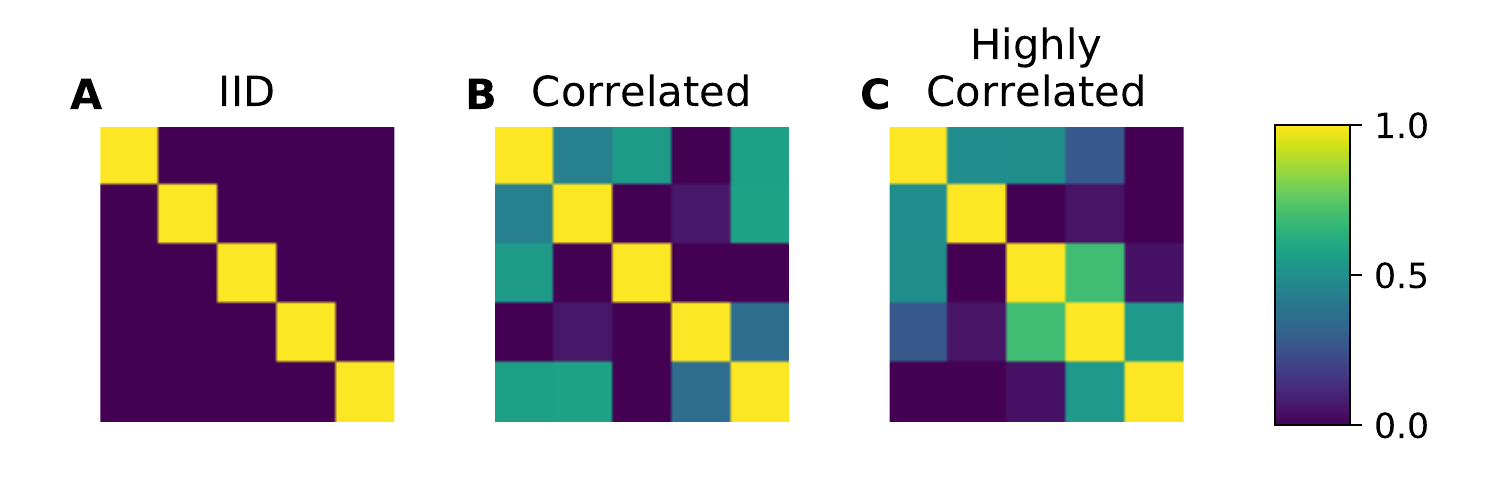}
    \caption{Covariances for different simulation settings in \tref{tab:sim_results}. Logistic regression accuracy improves with and without DAC, similiar to \tref{tab:logit_append}. Each point corresponds to classification accuracy for a different dataset. These datasets are the datasets in PMLB \cite{olson2017pmlb} for which a random forest outperforms logistic regression by at least five percent.}
    \label{fig:covs}
\end{figure}

\subsection*{Finding conditional expectations in real datasets}

It is difficult to validate that DAC curves are accurate as there is no groundtruth for what importance should be. As a surrogate, we use the conditional expectation curve calculated on an extremely large data sample. 

Specifically, we analyze the regression datasets in PMLB (a disjoint set of datasets as the classification datasets analyzed in \tref{tab:logit_append}). We fit the model and DAC/PDP curves to 10\% of the data. Then, we calculate the conditional expectation curve of the predictions of the model on the other 90\% of the data. \tref{tab:real_data_results} shows the results for the mean squared error between the conditional expectation curve and the DAC/PDP curves. The DAC curve incurs a substantially smaller MSE, suggesting that even with limited data, it is able to estimate the conditional expectations contained in the model.

\begin{table}[H]
  \caption{Mean squared error between importance curve (on a small data sample) and conditional expectation curve (on a large data sample). Averaged over multiple features on multiple datasets (9 total points). Error is standard error of the mean.}
  \label{tab:real_data_results}
  \centering
  \begin{tabular}{ccc}
    \toprule

    & \textbf{DAC MSE} & PDP MSE \\ 
    \midrule
    
    & 11.843 $\pm$ 3.948 &     21.175 $\pm$ 7.058 \\

    \bottomrule
  \end{tabular}
\end{table}

\begin{figure}[H]
    \centering
    \includegraphics[width=0.5\textwidth]{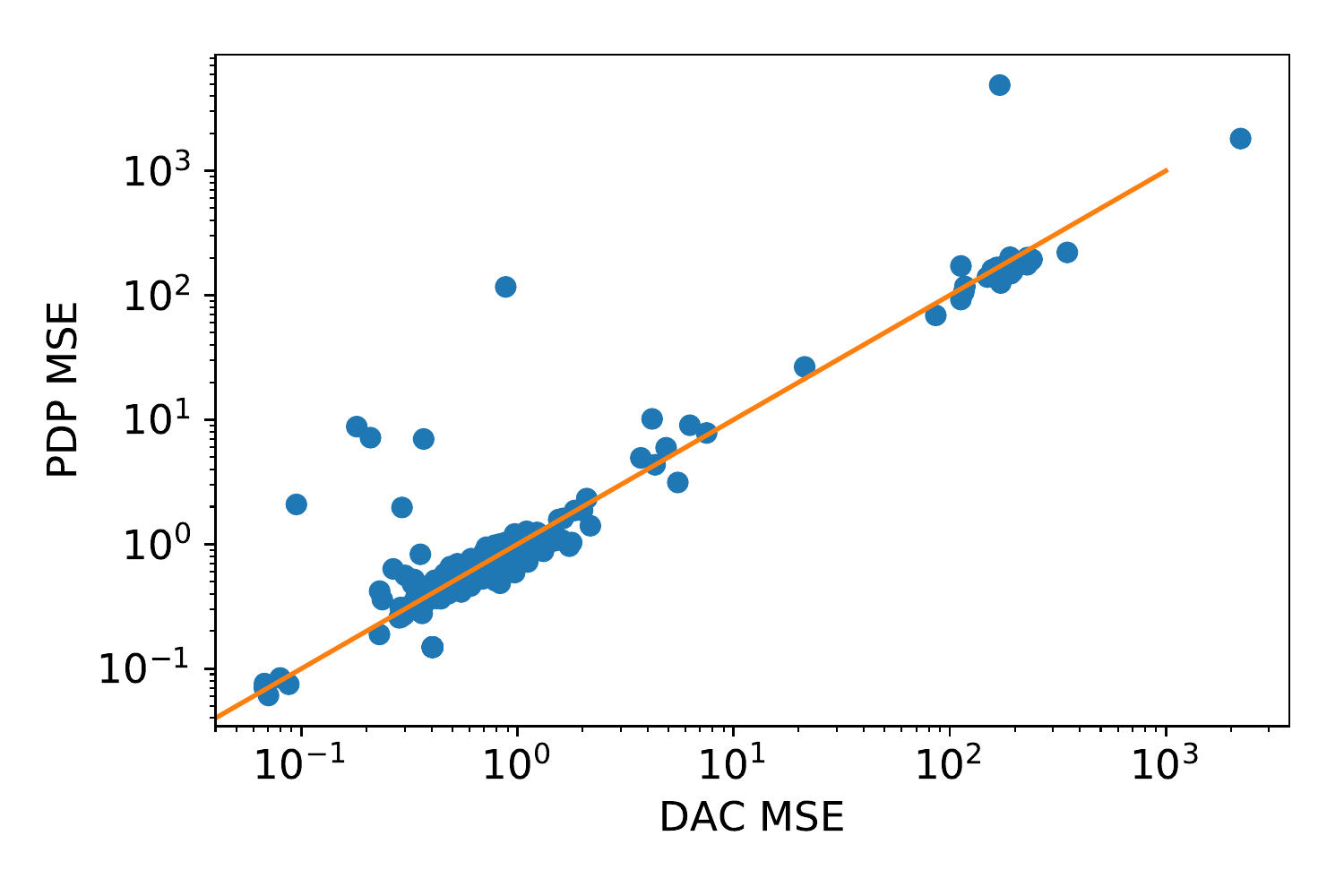}
    \caption{MSE for DAC is generally lower than MSE for PDP. Each point corresponds to MSE for a different dataset (1,521 points total). These datasets are the regression datasets in PMLB \cite{olson2017pmlb} for which a random forest outperforms logistic regression by at least five percent. Excluded a few points for which the MSE is above $10^6$, as these points are poorly fit by the random forest.}
    \label{fig:pmlb_reg}
\end{figure}